\documentclass[11pt,a4paper]{article}
\usepackage{times,latexsym}
\usepackage{url}
\usepackage[T1]{fontenc}

\usepackage[acceptedWithA]{tacl2021v1}
\usepackage{graphicx} 
\usepackage{xspace,mfirstuc,tabulary}

\newif\iftaclinstructions
\taclinstructionsfalse 
\iftaclinstructions
\renewcommand{\confidential}{}
\renewcommand{\anonsubtext}{(No author info supplied here, for consistency with
TACL-submission anonymization requirements)}
\newcommand{\instr}
\fi

\iftaclpubformat 

\else

\fi

\hypersetup{
           breaklinks=true,   
           colorlinks=true,   
           pdfusetitle=true,  
}

\title{Semantics of Multiword Expressions in Transformer-Based Models:\\ A Survey}

\author{
  Filip Miletić
  \and
  Sabine Schulte im Walde
  \\
  Institute for Natural Language Processing, University of Stuttgart, Germany
  \\
  \texttt{\{filip.miletic, schulte\}@ims.uni-stuttgart.de}
}

\date{}

\begin{document}
\maketitle

\begin{abstract}
Multiword expressions (MWEs) are composed of multiple words and exhibit variable degrees of compositionality.
As such, their meanings are notoriously difficult to model, and it is unclear to what extent this issue affects transformer architectures.
Addressing this gap, we provide the first in-depth survey of MWE processing with transformer models.
We overall find that they capture MWE semantics inconsistently, as shown by reliance on surface patterns and memorized information.
MWE meaning is also strongly localized, predominantly in early layers of the architecture.
Representations benefit from specific linguistic properties, such as lower semantic idiosyncrasy and ambiguity of target expressions.
Our findings overall question the ability of transformer models to robustly capture fine-grained semantics.
Furthermore, we highlight the need for more directly comparable evaluation setups.
\end{abstract}

\section{Introduction}
Multiword expressions (MWEs) -- such as noun compounds (e.g.\ \textit{jet lag}), particle verbs (e.g.\ \textit{take off}), and idioms (e.g.\ \textit{on the fly}) -- are composed of multiple words and exhibit semantic idiosyncrasy, i.e.\ their overall meaning cannot be directly predicted from the meanings of their constituents. 
They are ubiquitous, affect various applications, and as such have been extensively addressed in NLP research \citep{sag2002multiword, baldwin2010multiword}.
Although the widely used transformer-based language models have been  analyzed regarding their ability to represent various types of linguistic knowledge, we still lack consolidated insights into their processing of MWE semantics.
The present survey provides a critical overview of existing work on this issue.

By definition, the meaning of a MWE is distributed over multiple constituents.%
\footnote{This also applies to closed compounds realized as a single orthographic unit (e.g.\ \textit{flashback}) which comprises clearly identifiable constituents (\textit{flash} and \textit{back}).}
In some cases, these can be separated by intervening material (e.g.\ \textit{\underline{turn} the volume \underline{\smash{up}}}).
For any model, it is more challenging to capture the meaning of multiple lexical elements than that of a single word.
The overall meaning may further be compositional to various degrees, i.e.\ similar to the sum of the component parts (e.g.\ \textit{climate change}) or unrelated to it (e.g.\ \textit{silver bullet}).
Moreover, the same expression may be interpreted more or less compositionally depending on its context (e.g.\ \textit{kick the bucket}).

Transformer models are assumed to easily overcome some of these challenges since their meaning representations are inherently contextual and successful on a wide array of semantic tasks \citep{devlin2019bert, brown2020language}.
But does this imply a robust ability to represent MWE meanings?
The answer to this question is not trivial, as such an ability requires the models to systematically
(i)~capture the semantic contributions of multiple tokens, potentially including figurative and rare meanings; (ii)~weigh them based on variable degrees of compositionality; and (iii)~further specify the interpretation in context.

In order to establish the current state of knowledge on this issue, we provide the first in-depth survey of the fast-growing body of work on MWE representations in transformer models.
We aim to understand the extent to which different pretrained and optimized models are able to capture MWE semantics, as well as whether they are affected by representational or linguistic factors.
Most studies focus on a single expression type (e.g.\ noun compounds), task (e.g.\ machine translation), or related wider issue (e.g.\ phrase representations).
We broaden the perspective by systematizing insights from often disjoint strands of research, and identify priorities for future work.
Our findings more generally highlight that transformer models represent complex linguistic knowledge inconsistently.

We first provide general background on the surveyed MWE approaches (§\ref{sec:approaches}) and then zoom into transformer models.
We explore whether they inherently capture MWE information (§\ref{sec:model-properties}),
where it is localized (§\ref{sec:repres-info}),
and whether it is affected by linguistic properties (§\ref{sec:ling-properties}).
We conclude with a summary and directions of future work (§\ref{sec:conclusion}).

\section{Survey overview}
\label{sec:approaches}
This section motivates our survey by providing background on lexical semantic representations in transformer models, and then summarizing the most frequent implementations and tasks.

\subsection{Lexical semantics in transformer models}
\label{sec:semantics-transformers}
Language models based on the transformer architecture process information through a series of layers relying on the multi-head attention mechanism, which weighs each token in a sequence based on its similarity to the other tokens \citep{vaswani2017attention}.
As a sequence progresses through the layers, the representations become gradually more contextualized \citep{ethayarajh2019how}
and ultimately capture lexical semantic properties such as word senses \citep{wiedemann2019does}.
This mechanism should benefit MWE processing by distributing contextually provided semantic information over multiple tokens, but it is unclear where the models encode MWE information and to what extent.

Different types of linguistic information are not represented in the same way in the transformer architecture \citep{rogers2020primer}.
It has been suggested that surface features are localized in the lower, syntactic features in the middle, and semantic features in the higher layers \citep{jawahar2019what}.
But while higher layers do encode word senses \citep{coenen2019visualizing}, type-level lexical information is better accessed in lower layers \citep{vulic2020probing}.
Moreover, transformer models are affected by spurious effects of sequence position \citep{mickus2020what},
which questions the general robustness of their semantic representations.
The models also struggle with logical phenomena such as negation \citep{ettinger2020what},
and are not strongly dependent on word order but learn higher-level distributional patterns \citep{sinha2021masked}.
These observations indicate difficulties in capturing contextually provided meanings of multiple tokens -- a key property required for MWE semantics.
The mechanisms behind these issues may affect MWE representations, both in terms of overall reliability and features specific to this type of expression.

\subsection{Summary of surveyed approaches}
Our survey aims to bring together insights into how MWE meanings are represented in transformer models. 
We mainly draw on intrinsic evaluations targeting MWEs, conducted both on pretrained and optimized transformer models.
We also consider work on downstream tasks, but only if it has clear implications for model behavior.
We purposefully include papers using a variety of models, datasets, and evaluation strategies, whose results may not be directly comparable. However, broad coverage enables us to highlight different perspectives and remaining gaps in the literature.

We analyze the surveyed papers with respect to three lines of questions.
(i)~In general terms, can transformer representations capture MWE semantics, can they be optimized to more robustly represent these meanings, and can they generalize to unseen expressions?
(ii)~In terms of localization, which layers, modeled tokens, and contextual elements carry relevant representational information?
(iii)~Do any linguistic properties of MWEs affect the quality of their representations?

\paragraph{Models.}
The surveyed papers predominantly use transformer models with an encoder-only architecture.
These include BERT and its multilingual version mBERT, trained on the masked language modeling (MLM) and next sentence prediction tasks 
\citep{devlin2019bert};
RoBERTa, which introduced an optimized training procedure \citep{liu2019roberta};
its multilingual version XLM-R \citep{conneau2020xlmr};
computationally efficient derivatives ALBERT \citep{lan2020albert} and DistilBERT \citep{sanh2019distilbert};
and SBERT, optimized for sentence representations \citep{reimers2019sentencebert}.
Further papers use encoder-decoder architectures such as DeBERTa \citep{he2021deberta}, BART \citep{lewis2020bart}, and T5 \citep{raffel2020exploring};
and autoregressive models, in particular XLNet \citep{yang2019xlnet} and  GPT \citep{radford2019language, brown2020language}.

\begin{table*}[t]
    \centering
    \scalebox{0.87}{%
    \begin{tabular}{llrlll}
        \hline \hline
        \textbf{Phrases}    && \multicolumn{2}{l}{Similarity to other phrases} && Attributes \\
        \cline{1-1} \cline{3-4} \cline{6-6}
        direct link	        && \multicolumn{1}{r}{0.328} & access service 
                            && \textsc{immediacy}  \\
        formal education    && \multicolumn{1}{r}{0.508} & school board
                            && \textsc{formality} \\
        common man          && \multicolumn{1}{r}{0.672} & average person 
                            && \textsc{commonness} \\
        \hline \hline
        \textbf{Compounds}  && Compos.           & Synonyms       && Relations \\
        \cline{1-1} \cline{3-4} \cline{6-6}
        fairy tale          && 1.9 \small{± 1.3} & fable          && \textsc{about}  \\
        insurance policy    && 4.4 \small{± 0.9} & insurance plan && \textsc{about} \\
        birth rate	        && 4.7 \small{± 0.5} & fertility rate && \textsc{have} \\
        \hline \hline
        \textbf{Idioms} && \multicolumn{2}{l}{Literal occurrence} 
                        && Idiomatic occurrence \\
        \cline{1-1} \cline{3-4} \cline{6-6}
        in light of     && \multicolumn{2}{l}{\small in the light of a bedside lamp} 
                        && \small{in the light of this success} \\
        on the cards    && \multicolumn{2}{l}{\small read whatever is written on the card}
                        && \small{a decisive victory was on the cards} \\
        open one's eyes && \multicolumn{2}{l}{\small I opened my eyes and looked up}
                        && \small{it opened my eyes to the plight}  \\
        \hline \hline
    \end{tabular}}
    \caption{Example MWEs with common gold standard information, sampled from datasets on phrase similarity \citep{asaadi2019big} and attributes \citep{hartung2015distributional}; compound compositionality and synonyms \citep{cordeiro2019unsupervised} as well as semantic relations \citep{oseaghdha2007cooccurrence}; and idiomaticity \citep{haagsma2020magpie}.}
    \label{tab:mwe_examples}
\end{table*}

\paragraph{Tasks and datasets.}
The notion of MWE subsumes varied linguistic phenomena and related tasks.
We briefly review the tasks and datasets that are most often used in the surveyed literature \citep[for more extensive categories and definitions, see e.g.][]{baldwin2010multiword, constant2017multiword}.
Example expressions and different types of gold standard information are presented in Table~\ref{tab:mwe_examples}.

On the most general level, we look into representations of \textbf{phrases}.
These are groups of words which function as syntactic units and whose overall meanings are therefore derived from multiple lexical elements.
Studies adopting this general focus represent the  meanings expressed using specific syntactic patterns, e.g.\ subject--verb--object structures such as \textit{child--read--book}.
Evaluation datasets target phrase similarity and paraphrase or attribute detection \citep{mitchell2008vectorbased, hartung2015distributional, pavlick2015ppdb, asaadi2019big, strakatova2020all, pham2023pic}.

On a more specific level, we examine \textbf{noun compounds} (e.g.\ \textit{gold mine}).
Syntactically, these are also phrases -- with at least one modifier and a nominal head -- and they are analyzed with a focus on semantic idiosyncrasy.
Tasks include predicting the degree of compositionality, i.e.\ the semantic relatedness of the constituents to the overall meaning; and predicting the meaning of the compound and evaluating it by detecting synonyms, paraphrases, or semantic relations.
These tasks rely on a wide range of datasets \citep{biemann2011distributional, reddy2011empirical, hendrickx2013semeval2013, juhasz2015database, levin2019systematicity, cordeiro2019unsupervised, pinter2020nytwit};
for a recent analysis, see \citet{SchulteImWalde:23}.

\textbf{Idioms} are structurally diverse phrases with conventionalized meanings which cannot be deduced from their constituents (e.g.\ \textit{spill the beans}).
Their literal vs.\ idiomatic interpretation often depends on context, so they are also referred to as potentially idiomatic expressions (PIEs).
A standard evaluation is idiomaticity classification on the sentence or token level
\citep{cook2008vnctokens, hashimoto2008construction, savary2017parseme, aharodnik2018designing, moussallem2018lidioms, haagsma2020magpie, saxena2020epie}.

Some papers define their MWE tasks as figurative language or metaphoricity detection.
Since figurative language involves non-compositionality and contextual specification, this is largely a matter of perspective.
These studies also examine phrases and idioms from the cited or ad-hoc datasets, using tasks such as idiomaticity detection (e.g.\ is a phrase such as \textit{political storm} used idiomatically?)
and plausibility classification (e.g.\ given a text ending with an idiom, is a candidate continuation plausible?).

\section{General model properties}
\label{sec:model-properties}
In this section, we first assess if MWE meanings are inherently captured by off-the-shelf transformer models, and if these can be further optimized so as to improve MWE representations.
We then look into general mechanisms that may support this ability: recall of memorized information and generalization to unseen data.

\subsection{Off-the-shelf representations}
\label{sec:off-the-shelf}
We begin by asking if pretrained transformer models  capture MWE meanings without optimization for this type of expression.
We examine the extent of this ability based on tasks targeting different compositionality ranges and objectives: predicting an expression's meaning or its semantic properties, e.g.\ the degree of compositionality.

A model that encodes MWE semantics should fulfill the necessary (but not sufficient) requirement of \textbf{representing compositional phrase meaning}, i.e.\ the overall meaning that is  derived from the meanings of the constituents; this would indicate that the model can capture semantics beyond the level of individual tokens.
This issue has been evaluated on the task of predicting phrase similarity.
Focusing on English phrases, \citet{gamallo2021comparing} show that the cosine similarity between phrase-level SBERT embeddings  -- corresponding to averaging over tokens -- is positively correlated with human similarity ratings, reaching $\rho = 0.61$ for noun-verb-noun expressions.
In a subsequent study on Galician, their best method uses the contextualized embedding of the verb \citep[$\rho = 0.57$;][]{gamallo2022evaluating}.
These studies show that encoding a phrase's constituent words via transformer architectures can produce meaningful representations of the whole phrase.
This in turn indicates that attention-based contextualization effectively distributes (some part of) phrase-level meaning over the constituent tokens.

This tendency is further confirmed by the fact that phrase-level representations can be reconstructed from the representations of their constituents.
Even with a straightforward strategy such as vector addition, the mean cosine between the original and reconstructed \texttt{CLS} embeddings reaches $0.92$ for BERT and $> 0.99$ for RoBERTa and DeBERTa \citep{liu2022are}.
But successful reconstruction of phrase representations does not entail that they are underpinned by refined compositional processing.
If that was the case, they would not be affected by surface factors such as word overlap.
\citet{yu2020assessing} consider the special case of phrases with inverted constituents (e.g.\ \textit{law school} and \textit{school law}), corresponding to 12\% of phrase pairs they use.
The correlation between model-derived phrase similarity and human ratings drops from $\approx$$0.6$ on the full dataset to $\approx$$0.2$ on the inverted constituent subset, and indicates a strong effect of word overlap.

MWEs exhibit \textbf{variable degrees of compositionality}, which should be reflected by appropriate semantic representations.
This has been investigated by predicting the compositionality of noun compounds using features extracted from pretrained models.
\citet{nandakumar2019how} report positive Pearson's correlation with human compositionality ratings, ranging from $r=0.15$ to $0.60$ depending on the dataset and estimation strategy.
However, the best BERT results are systematically $\approx$$0.2$ points behind the strongest methods based on static word embeddings.
In a more extensive evaluation, \citet{garcia2021assessing} reach $\rho = 0.37$ on English and $0.26$ on Portuguese data, similarly lagging behind the SOTA on static word embeddings \citep[$0.73$ and $0.60$, respectively;][]{cordeiro2019unsupervised}.
These results question BERT's ability to capture compositionality similarly to humans, but they may be due to a suboptimal use of the information encoded in models which -- unlike static word embeddings used on this task -- do not learn dedicated MWE representations.

Subsequent work by \citet{miletic2023systematic} has shown that robust compositionality information \textit{can} be extracted from BERT, but it is not equally accessible across the model architecture.
The best results (reaching $\rho=0.71$) are obtained using embeddings from early layers and comparisons between compounds and their contexts.
More generally, \citet{shwartz2019still} have examined the potential of transformer representations to capture degrees of compositionality across a variety of tasks.
In their supervised classification setup, contextualized representations systematically outperform static word embeddings.
However, even implementations that are overall good at distinguishing degrees of compositionality struggle when exposed to more complex semantic mechanisms such as implicit meaning.
Accuracy is in the $80-90\%$ range on literality-related tasks, but it drops by $\approx$$30\%$ points on noun compound relations and adjective--noun attributes.

Another key issue is whether \textbf{non-compositional MWE meanings} are represented as such, i.e.\ more similarly to an independent linguistic unit than a sum of component parts.
One line of evidence questioning this ability comes from patterns of similarity between non-compositional expressions.
\citet{zeng2022getting} extract mean-pooled idiom embeddings from BART and find that they cluster together based on surface or syntactic similarity rather than figurative meaning.
\citet{garcia2021probing} compare contextualized embeddings of compounds and their synonyms.
They assume that their similarity should \textit{not} correlate with compositionality ratings if compound meanings are represented well across compositionality ranges.
However, they find moderate-to-strong correlations across models for both English and Portuguese. 
This indicates that non-compositional compounds are further away from their synonyms in the vector space and to that extent are represented less well than compositional compounds.

A more nuanced picture emerges from attention flows in neural machine translation, as examined by \citet{dankers2022can} on PIEs.
In figurative contexts, PIEs exhibit increased self-attention within the expression and reduced interaction with the surrounding context.
This suggests they are grouped together more strongly, i.e.\ processed similarly to a standalone linguistic unit.
On the decoder side, this is echoed by lower cross-attention between figurative translations and source PIEs.
However, when encoded information is progressively removed through amnesic probing, the model reverts to compositional translations.
This brittleness highlights the challenging nature of figurative translations.

\textbf{Summary.} Moderately strong results across tasks of different complexity indicate that pretrained models capture MWE semantics, but do so inconsistently.
This is further shown by their reliance on surface patterns such as word overlap, strong localization of relevant information, and comparatively lower quality of non-compositional meaning representations.

\subsection{Optimized representations}
The shortcomings of MWE representations raised in the previous section may be addressed using different approaches.
We discuss span representations, which are optimized to capture the meaning distributed over multiple tokens;
task-specific fine-tuning or adapter-tuning, with training strategies that target properties typical of MWE semantics;
enhancing models with linguistic knowledge, such as explicit information on potential interpretations of MWEs;
and training dedicated neural architectures, which rely on greater model complexity to improve MWE representations.

General-purpose models that are optimized to \textbf{represent spans of text}
should better capture the meaning of multiple tokens and, by extension, MWEs.
SBERT produces sentence-level embeddings following fine-tuning on NLI data with a siamese architecture \citep{reimers2019sentencebert}.
SpanBERT is pretrained by masking contiguous spans of text instead of individual tokens, and uses span boundary representations to encode span content \citep{joshi2020spanbert}.
PhraseBERT fine-tunes BERT using contrastive learning over positive and negative examples of paraphrases and of contexts \citep{wang2021phrasebert}.
Among these, SpanBERT obtains the best results on in-context phrase similarity; across evaluations setups, its strongest improvement over BERT is $2.2$ accuracy points \citep{pham2023pic}.
For type-level phrase similarity, better performance can be obtained by aggregating the similarities of multiple pairs of occurrences at inference time. 
The improvement over BERT stands at $8.6$ to $28.7$ accuracy points depending on the dataset \citep{cohen2022mcphrasy}.

Turning to representations more specifically targeting MWEs, different \textbf{fine-tuning approaches} have been proposed.
Following the findings of \citet{yu2020assessing} regarding strong effects of surface patterns on phrase representation, \citet{yu2021interplay} fine-tune models to avoid this effect, for example by predicting if two sentences with high lexical overlap are paraphrases or not.
This only leads to minor localized improvements of phrase representations which do not reduce the reliance on word overlap.
However, fine-tuning has a stronger effect in other setups.
\citet{liu2022testing} evaluate figurative language interpretation using a Winograd-style task targeting novel metaphors.
Their strongest model is RoBERTa fine-tuned with a contrastive objective, reaching $90.3\%$ accuracy, within $5$ points of human performance.
This is an improvement of $24.1$ points on the zero-shot setup.

A computationally leaner approach consists in \textbf{learning an adapter}, as shown by \citet{zeng2022getting} on BART idiom embeddings.
They evaluate different adapters, with learning objectives that include reconstructing corrupted idiomatic sentences and increasing the similarity between the embeddings of idioms and their dictionary definitions.
They obtain clear improvements across evaluations, e.g.\ accuracy on idiom span detection increases from $50.8$ to $76.3$.
Fine-tuning the full model performs similarly to the directly comparable adapter; it is outperformed by the best adapter variants, trained on additional objectives.
An acknowledged limitation is difficulty in generalizing to unseen idioms, stemming from the use of external linguistic knowledge as a supervision signal.

Fine-tuned models can be somewhat further improved using \textbf{external linguistic knowledge}.
\citet{chakrabarty2022it} evaluate models on a binary classification task targeting plausible continuations of narrative texts whose last sentence contains an idiom or a simile.
Their strongest zero-shot approach obtains a relatively high accuracy of $67.7\%$ on unseen idioms.
It is strongly outperformed by a model with task-specific fine-tuning ($82.0\%$),
but a further improvement of $1.5$ accuracy points is obtained by providing knowledge of the literal meaning of idiom constituents.

Targeted improvements have also been obtained using \textbf{more complex dedicated architectures}.
\citet{zeng2021idiomatic} approach idiomaticity detection by using the attention flow mechanism \citep{seo2017bidirectional} to fuse BERT-derived representations with static word, character, and POS embeddings. 
On sentence-level idiomaticity classification across multiple datasets, this method performs similarly or worse than a standard implementation with a linear layer on top of BERT.
However, on a stricter accuracy measure -- where each token in a sequence is required to be accurately classified for idiomaticity -- it outperforms the standard approach by a margin of $20-30$ points.
Focusing on Chinese idiom recommendation in a cloze task, \citet{tan2020bertbased} use the \texttt{MASK} embedding to retrieve the correct idiom with a $79.8\%$ accuracy, 
whereas their ‘‘dual embedding'' approach -- capturing both the immediate context and the broader textual passage -- leads to an improvement of $2.6$ points.
\citet{tan2021bertbased} subsequently propose a dedicated BERT model pretrained on the MLM task by only masking idioms, and then fine-tuned for multiple-choice recommendation.
It reaches $86.3\%$ accuracy, within a point of human performance.
These results overall indicate that dedicated architectures can achieve excellent performance on some tasks.

\textbf{Summary.} Different strategies improve MWE representations, with gains over pretrained models varying from marginal to dramatic.
The viability of these methods should be carefully weighed against the expected improvements, especially for computationally expensive systems.
However, this requirement remains difficult to fulfill: optimization strategies differ widely in terms of generality (targeting any sequence of tokens vs.\ a specific type of MWEs), evaluation complexity, generalizability to unseen data, and underlying architectures. More comprehensive evaluations enabling direct comparisons are a priority for future work.

\subsection{Memorization and generalization}
\label{sec:memorization}
Building on our earlier finding that transformer models encode some knowledge of MWEs, we now look into general mechanisms that enable it.
We examine the reliance on memorized information and the complementary generalization ability.

Transformer models seem to process MWEs largely based on the \textbf{recall of memorized expressions} rather than a sophisticated meaning processing mechanism.
When interpreting novel compounds, GPT-3 can provide human-like explanations but it seems to draw on memorized token distributions rather than reason about the underlying conceptual categories \citep{li2022systematicity}.
Noun compound paraphrases generated by GPT-3 substantially overlap with web content which likely constitutes its training data; this trend is stronger for existing than for novel compounds.
The acceptability of generated paraphrases is lower for novel compounds, which may partly reflect the lack of memorized information \citep{coil2023chocolate}.

Using the task of predicting the final token of an idiom given its preceding tokens, \citet{haviv2023understanding} report that GPT-2 has memorized $45-48\%$, and BERT $28-38\%$ of expressions from their set of $\approx$$800$ items, with higher scores in larger model variants.
They further show that memorized information is retrieved in two distinct stages:
(i)~early layers promote a decrease in the rank of the target token, bringing it closer to the top of the candidate token set;
(ii)~later layers promote an increase in its probability.
Memorized idioms undergo a slower first stage, i.e. target completions reach the top of the distribution in comparatively later layers, potentially due to the processing of the full input and not only the local context.
They also exhibit a more pronounced second stage, with final probabilities around three times higher compared to non-memorized idioms; this is consistent with a smaller set of plausible completions.
These findings have important implications for methods that represent MWEs using representations from a specific layer, as the optimal choice may depend on the degree of memorization of the target expression (i.e. memorized expressions may be better represented in comparatively later layers; for further discussion of layers, see §\ref{sec:layers}).

Reliance on memorized information is desirable in some settings -- for example, when generating highly conventionalized expressions such as idioms -- but it may hinder \textbf{generalization ability} on other tasks.
\citet{falk2021automatic} evaluate BERT on attribute selection for German adjective--noun phrases, framed as multiclass classification (e.g.\ \textit{schlauer Junge} ‘smart boy' has the attribute label \textit{intelligence}).
Compared to evaluation on unseen data, performance is stronger when train and validation/test sets have a partial lexical overlap, i.e.\ the same set of heads or of modifiers.
Depending on the dataset variant, this can lead to an improvement of up to $0.26$ F1 with modifier overlap.
The stronger effect for modifiers (here, adjectives) is consistent with their central role in this task.

These results are echoed by a more general trend that model performance tends to follow: seen data $\gg$ unseen data $>$ cross-lingual data.
\citet{fakharian2021contextualized} evaluate a range of models on PIE idiomaticity classification in English and Russian.
The general trend is illustrated by mBERT with task-specific fine-tuning on English.
It achieves $83.8\%$ accuracy on seen English data, $74.3\%$ on unseen English data, and $72.4\%$ on Russian data.
A similar drop is observed across transformer models, but they remain clearly above baselines, indicating a non-negligible ability to generalize.
The same task is investigated for Slovene by \citet{skvorc2022mice}.
They compare mBERT, pretrained on 104 languages, including Slovene; and CroSloEngual-BERT, pretrained only on Croatian, Slovene and English.
Looking at sentence-level classification, mBERT is weaker on seen idioms ($0.91$ vs.\ $0.95$ F1) but better on unseen idioms ($0.90$ vs.\ $0.84$).
This suggests that it is stronger at generalizing -- perhaps due to pretraining on multiple languages related to Slovene -- as further confirmed by above-chance cross-lingual performance on Croatian ($0.90$) and Polish ($0.70$).

Models can also be optimized for generalization.
From a dataset perspective, a BERT-based idiomaticity classifier reaches generalizability faster if the idioms to which it is exposed during training are ordered by decreasing contribution to model performance.
The contribution is determined by an idiom's Shapley value, estimated as the difference between the average performance of multiple models which do vs.\ do not include a given idiom in training data \citep{nedumpozhimana2022shapley}.
From an architecture perspective, the previously discussed use of attention flow to fuse contextualized and static idiom representations is especially beneficial for generalization to unseen idioms and to other domains \citep{zeng2021idiomatic}.
This finding -- contrasted by competitive performance of standard architectures on seen data -- once again indicates that more complex systems are particularly useful in challenging classification scenarios.

\textbf{Summary.} Transformer-based MWE representations strongly rely on memorized information, as observed when generating subparts or paraphrases of target expressions.
Novel (non-memorized) expressions yield lower-quality generations and are processed in earlier layers, indicating dominance of the local context.
Models generalize to unseen and cross-lingual data with a performance drop in tasks of variable semantic complexity; this can be alleviated by targeted optimization.

\section{Impact of representational information}
\label{sec:repres-info}
Shifting the focus from general insights into MWE semantics captured by different transformer models,
we now adopt a finer-grained perspective and examine how MWE representations are impacted by structural factors, i.e.\ model and input properties which directly affect the representational information extracted from a given transformer architecture.
We address three such factors: transformer layers, tokens within the sequence, and the context surrounding the target expression.

\subsection{Layers}
\label{sec:layers}
As previously noted (§\ref{sec:semantics-transformers}), representations from different transformer layers do not capture the same range of linguistic information.
We explore the effect that this has on MWE representations by reviewing standard layer choices, their variable effects on performance, interactions with model and linguistic properties, and potential explanations.

When \textbf{selecting the layers} to represent MWE meanings, a common choice is the last layer, both as input to a classifier \citep{nedumpozhimana2021finding, nedumpozhimana2022shapley}
and as a standalone representation, often constituting a baseline for an optimized model \citep{wang2021phrasebert, pham2023pic}.
Representations pooled over the last four layers have also been used with variable degrees of success \citep{gamallo2021comparing, garcia2021assessing}.
Other methods learn a scalar mix of layers \citep{falk2021automatic};
one study has found that a balanced mix of top and bottom layers tends to outperform the individual use of the last layer across tasks \citep{shwartz2019still}.

The impact of layer choice has been assessed on a range of tasks.
Most surveyed papers report \textbf{better performance in lower layers}; recall that these are the least contextualized representations, assumed to capture surface linguistic features.
\citet{brglez2023dispersing} evaluates metaphoricity prediction on 24 Slovene noun phrases, with the best results in the input embedding layer 0.
The cosine similarity between consitutents is initially higher in literal than metaphorical examples, as expected, but this difference diminishes over the layers.
\citet{miletic2023systematic} predict the degrees of compositionality of 280 English noun compounds, and similarly find the most consistent performance in the low-to-mid range of layers, with the single best result on layer~1.
They experiment with pooling contiguous layers, but this penalizes performance.
On PIE idiomaticity classification, \citet{tan2021does} find that performance stabilizes around layer~4 and generally peaks in the mid-range, indicating that several rounds of contextualization are sufficient for this task.
Predicting compositionality or idiomaticity may come down to identifying discrepancies between the target expression and its context, which would support the preference for the less contextualized, lower layers. But similar results have been reported on tasks requiring comparisons of multiple target expressions. For instance,
\citet{burdick2022using} evaluate paraphrase similarity on over 25k phrase pairs and obtain the best individual result with layer~1.

Contrasting this trend, better performance in higher layers has occasionally been reported when predicting PIE idiomaticity \citep{fakharian2021contextualized} and the semantic transparency of closed compounds \citep{buijtelaar2023psycholinguistic}.
These differences may be explained by the fact that information encoded by different layers is affected by \textbf{interactions with other parameters}, such as the choice of the pretrained model architecture and of the target token (e.g.\ modeled in isolation or in sentence context).
Detailed evidence of this trend comes from evaluations of phrase similarity by \citet{yu2020assessing}.
They experiment with modeling the target expressions without additional context, and observe the best performance in earlier layers for RoBERTa, XLM-R, and XLNet; middle layers in BERT; and later layers in DistilBERT.
By contrast, when sentence context is included, layers in the mid-range are generally strongest for all models.
Moreover, the \texttt{CLS} embedding improves in performance as layers progress, pointing to distinct processing of the information it captures.

Layer-level information is also affected by the \textbf{linguistic properties} of the modeled expressions.
Focusing on the task of paraphrase identification, \citet{tan2021does} report an effect of the degree of idiomaticity:
when both the target expression and the paraphrase are non-idiomatic, performance is strongest at layer~0 and decreases afterwards;
when the target expression is idiomatic (and the paraphrase is either idiomatic or not), performance is relatively stable across the layers.
Similarly, \citet{burdick2022using} estimate paraphrase similarity.
Within a pair of paraphrases used in the same context, the same words become less similar, and different words more similar, as layers progress; this confirms that later layers capture more contextual information.
These findings are echoed by the processing of PIEs in machine translation, examined by \citet{dankers2022can}. 
As layers progress, figuratively used PIEs become less similar to their representations in the preceding layer, compared to their literal counterparts, suggesting a stronger effect of contextualization.

Potential explanations for these trends are provided by the models' \textbf{structural features}.
\citet{aoyama2022probeless} show that different types of MWE information do not follow the same distribution over layers.
When predicting a MWE token, the model tends to rely on lower layers compared to all tokens, perhaps due to a smaller set of potential candidates.
When predicting POS tags, it tends to rely on higher layers compared to all tokens, which may be related to the usefulness of semantic information in resolving POS sequences typical of MWEs.
\citet{espinosaanke2021evaluating} note the effects of anisotropy in BERT, i.e.\ the tendency for embeddings to concentrate in a narrow cone.
On collocate categorization, this leads to overlaps between antonymic collocates -- which should ideally be distant in the vector space -- with the authors questioning whether the model has inherent knowledge to resolve their task.
\citet{klubicka2023idioms} show that, within a given layer, idiomaticity is mostly encoded in vector dimensions rather than the norm, and is somewhat more accessible in the first half of the dimensions. This confirms that linguistic information of interest is not equally distributed over an embedding.

\textbf{Summary.} Later transformer layers are often used to represent MWEs, but lower layers are generally better when predicting both an expression's meaning and properties such as compositionality.
This suggests that MWE semantics are best captured by weakly to moderately contextualized representations, highlighting in turn the relevance of type-level lexical information.
This trend is affected by model properties as well as key linguistic features, e.g.\ figurative and idiomatic expressions benefit from stronger contextualization.
But this mirrors the patterns noted for memorized expressions (§\ref{sec:memorization}); future work should therefore analyze interactions between idiomaticity and memorization.
More immediately, the observed patterns indicate that layer choice should be carefully tuned.

\subsection{Tokens}
After inputting a MWE into a transformer model, embeddings of multiple tokens of interest may be used to represent it.
We first address the standard choices and their effects on performance before zooming into attention-based contextualization.
We then discuss two implementation issues: the \texttt{CLS} token and subword fragmentation.%
\footnote{In what follows, \textit{constituent} denotes an expression's constituent word without implication for syntactic properties.}

When \textbf{selecting the modeled token}, frequent choices include the embedding of a constituent, modeled as part of the full expression and thereby contextualized relative to the other constituents \citep{brglez2023dispersing};
a phrase representation obtained by pooling the constituent embeddings \citep{pham2023pic};
the embedding of the \texttt{MASK} token, replacing the target expression in a sequence \citep{tan2020bertbased};
and the embedding of the \texttt{CLS} token, corresponding to the sequence containing the target expression \citep{fakharian2021contextualized}.
Unsurprisingly, embeddings of different tokens do not capture the same information; they may in fact be complementary.
On English and Japanese idiom token classification, \citet{takahashi2022leveraging} gain $\approx$$0.025$ accuracy points by concatenating contextualized, out-of-context, and \texttt{MASK} representations of constituents; this is opposed to using the contextualized embeddings of constituents.

The \textbf{effect of token choice} has been examined on several tasks.
On phrase similarity, \citet{yu2020assessing} show that phrase representations averaged over the constituents obtain better results than alternatives, such as the embedding of the head or of the full sequence.
Using a similar approach and task on English, \citet{gamallo2021comparing} report better results with phrase representations; 
on Galician, \citet{gamallo2022evaluating} obtain a slight improvement with constituent embeddings ($\rho$ increase of $0.03$ compared to phrase embeddings).
This may be explained by language-specific patterns or by implementation differences (phrase embeddings obtained using English SBERT vs.\ simple mean pooling from Galician BERT variants).
As for compound compositionality prediction, \citet{miletic2023systematic} obtain the best results by comparing the embedding of the constituent of interest -- entire compound to predict compound-level compositionality, and the head or modifier to predict their respective contributions to compound meaning -- with a pooled embedding of the surrounding sentence context.

Even where post-hoc pooling is required to represent a target item, attention-based \textbf{contextualization has beneficial effects}.
On the task of adjective attribute classification, \citet{falk2021automatic} find that phrase embeddings generally outperform constituent embeddings.
However, the modifier (i.e.\ adjective) embeddings are at most slightly behind, indicating that they carry most task-relevant information which is further distributed through contextualization.
In order to predict compound compositionality, \citet{garcia2021assessing} compare compound embeddings obtained within a sentence and out-of-context. 
For the out-of-context setting, they apply pooling over representations of constituents obtained by feeding the model (i)~with the entire compound; (ii)~with each constituent individually. 
They obtain better results with (i) ($\rho=0.37$ vs.\ $0.16$), showing that contextualization via self-attention provides a stronger contribution than a simple composition operation.

In addition to performing well on specific tasks, \textbf{pooled representations compared to standalone embeddings} capture some meaningful linguistic information.
\citet{nandakumar2019how} predict compound compositionality by comparing constituent and compound BERT embeddings, obtaining a positive correlation with human ratings (up to $r=0.38$).
\citet{garcia2021probing} likewise assess the similarity of a mean-pooled compound representation and that of only one constituent.
They obtain generally high cosine scores ($\approx$$0.8$), showing that the representations are closely similar but not identical; and a weak to moderate correlation with compositionality ratings (up to $\rho=0.45$).

In terms of specific tokens, \textbf{the \texttt{CLS} token encodes clearly distinct information} relative to tokens corresponding to MWE constituents.
For example, its use on compound compositionality prediction leads to stark drops in performance compared to other tokens of interest \citep{miletic2023systematic}.
However, this trend may be partly model-specific.
On phrase similarity, \texttt{CLS} generally performs poorly except for DistilBERT, where it appears to encode a compositionality signal \citep{yu2020assessing}.
When reconstructing a phrase representation from its constituents, averaging over all tokens outperforms \texttt{CLS} for BERT, RoBERTa, and DeBERTa -- but GPT-2 performs better using the (roughly equivalent) sequence-final token \citep{liu2022are}.

A connected issue is \textbf{subword fragmentation}: if a word is not present in a model's vocabulary, it is tokenized into smaller fragments for which representations exist.
Standard solutions include averaging over the subword tokens \citep[e.g.][]{garcia2021assessing}
or using only the first \citep[e.g.][]{gamallo2021comparing}.
These solutions are not detrimental when subword fragmentation affects a small subset of target items \citep{miletic2023systematic},
but it can be widespread for specific structures.
Focusing on English closed compounds, \citet{pinter2020will} compare representations obtained by pooling subword-fragmented BERT embeddings vs.\ those that are first pre-tokenized into gold-standard constituents.
Similarity between the two types of pooled representations for a given compound is high overall (cosine reaching $\approx$$0.8-0.9$) but is affected by additional factors: it increases over layers, peaking at layer~11 of 12; and it is stronger for more semantically transparent items.
Put differently, subword-fragmented and linguistically motivated representations of constituents recover similar compound-level information, with benefits from attention-based contextualization in both cases.
\citet{jenkins2023split} analyze German (closed) compounds and find that pre-tokenization into constituents is beneficial for some evaluations, highlighting the relevance of the target task.

\textbf{Summary.} Multiple lines of evidence converge to indicate that MWEs are best represented by tokens corresponding to linguistic structures of interest, contextualized within the expression and pooled where necessary.
This is facilitated by self-attention, which distributes linguistic information over tokens.
The \texttt{CLS} token requires cautious implementation due to idiosyncrasies. Subword fragmentation is generally not detrimental.

\subsection{Contextual information}
Transformer models can represent sequences of variable lengths, so we now examine the consequences of modeling MWEs in isolation and in sentence context.
We show the benefits of broader context and variants of this information, and then look at how it interacts with model mechanisms.

There is a clear consensus that \textbf{contextual information is beneficial} for modeling MWEs.
Increasing the amount of linguistic context -- including any (rather than none) or including more (rather than some) -- improves performance on tasks including
phrase similarity estimation \citep{cohen2022mcphrasy},
idiom translation \citep{baziotis2023automatic},
metaphoricity prediction on noun-verb phrases \citep{brglez2023dispersing},
and compositionality prediction on open \citep{miletic2023systematic} and closed compounds \citep{buijtelaar2023psycholinguistic}.
Contextual information is at the core of some approaches, e.g.\ idiomaticity detection assuming semantic compatibility between literal MWEs and their context \citep{zeng2021idiomatic}.
Evidence disputing the usefulness of context is mostly limited to improvements that are strong overall but absent in a subset of settings, e.g.\ on phrase similarity \citep{pham2023pic}.

Different \textbf{variants of contextual information} have been proposed.
Representations of phrases -- including syntactic structures typical of compounds (noun and adjective phrases) and idioms (verb phrases) -- can be obtained through contrastive fine-tuning on paraphrases and further improved by extending the procedure to phrase contexts, i.e.\ fine-tuning on entire sentences in which phrases appear;
accuracy gains reach $8.9$ points on longer sequences \citep{wang2021phrasebert}.
Chinese idiom prediction is similarly improved by including paragraph-level context in addition to the target sentence \cite{tan2020bertbased}.
Compound compositionality prediction strongly benefits from modeling paraphrases in addition to compound occurrences, with $\rho$ increasing by $0.5$ points compared to only using the targets and their constituents \citep{nandakumar2019how}.
More generally, increasing the number of modeled instances per expression leads to an increase in performance; it levels off after $\approx$$100$ examples on phrase similarity \citep{cohen2022mcphrasy}.
On compound compositionality prediction, the improvement is strongest when shifting from 10 to 100 examples, and minor with a further shift to 1,000 examples \citep{miletic2023systematic}.

Linguistic context may \textbf{interact with representational properties} such as layers, underlying architectures, and modeled tokens.
On phrase similarity, contextual information improves results overall and reduces the impact of individual layers. 
When only the target expressions are modeled, performance tends to drop as layers progress; when the expressions are in sentence context, it is largely stable \citep{yu2020assessing}.
Model-specific patterns have been reported on retrieval of idioms vs.\ compositional phrases. 
Inclusion of context has no effect on BERT and T5; it reduces surprisal for GPT-2 variants but without altering the patterns relative to other settings \citep{rambelli2023are}.
Representations are also affected by word position, shown on pairs of paraphrases in the same context. Same words appearing in different positions are considerably less similar to one another, compared to both same and different words in the same position; this is stronger for larger changes in position \citep{burdick2022using}.

Models draw on \textbf{different sources of linguistic information} provided as input.
Collocate categorization improves when MLM predictions -- where the target expression is masked in a sentence -- are conditioned on the full non-masked sentence by concatenating it \citep{espinosaanke2021evaluating}.
In classification of compound semantic relations and adjective attributes, the strongest results are obtained using the target expression, in sentence context, together with a paraphrase; omitting any of the three elements reduces accuracy by up to $9.4$ points \citep{shwartz2019still}.
Similarly, probing experiments on idiomaticity classification indicate that BERT relies on information localized mainly in the idiomatic expression itself, but also in the surrounding context \citep{nedumpozhimana2021finding}.
This is indirectly echoed by better performance on individual items whose topic distribution is similar to that of the full dataset \citep{nedumpozhimana2022shapley}.

\textbf{Summary.} A wide array of experimental settings unequivocally show that any increase in contextual information enables better MWE representations. Linguistic context affects the behavior of model structures and is a beneficial source of information on multiple tasks -- including those which are not readily reduced to comparisons of target expressions and surrounding context.

\section{Impact of linguistic properties}
\label{sec:ling-properties}
MWE representations may be affected by properties of the target expressions themselves.
We now provide a breakdown of the reported effects.

Individual expressions vary in terms of their inherent \textbf{predictive properties}, as shown in work on the usefulness of individual idioms when training an idiomaticity classifier.
\citet{nedumpozhimana2022shapley} note a positive effect of informativeness, measured by training a classifier on one idiom and evaluating it on the full set of idioms; and ease of prediction, measured by training a classifier on the full set of idioms and evaluating it on one idiom.

Models are affected by the degree of \textbf{semantic idiosyncrasy}.
\citet{falk2021automatic} obtain better results for attribute selection on phrases with higher semantic transparency.
On idiomaticity detection, \citet{zeng2021idiomatic} report a slight gain ($\approx$$0.03$ F1) for expressions that are fixed rather than semi-fixed or syntactically flexible.
Non-idiomatic expressions are better represented in lower layers \citep[see~§\ref{sec:layers};][]{tan2021does}.

As for other semantic properties, lower \textbf{polysemy} is associated with better results on attribute selection \citep{falk2021automatic} and compositionality prediction \citep{miletic2023systematic},
while it does not affect word similarity across paraphrase pairs \citep{burdick2022using}.
More \textbf{concrete} words are assigned more weight when estimating constituent contributions to compound meaning \citep{buijtelaar2023psycholinguistic}.
The \textbf{type of semantic knowledge} affects metaphor interpretation, with an evaluation on 10k examples showing better results for object and visual commonsense metaphors (referencing common objects and their visual attributes) than for social and cultural commonsense metaphors (referencing human behavior and cultural norms) \citep{liu2022testing}.

Varied effects have been noted for \textbf{frequency}.
They may be partly model-specific, with low-frequency compositional phrases yielding higher surprisal in GPT-2 and T5, but not BERT \citep{rambelli2023are}.
Different noun compound analyses have reported that frequency has no effect \citep{buijtelaar2023psycholinguistic},
that low frequency is detrimental \citep{coil2023chocolate}
and that it is beneficial \citep{miletic2023systematic}.
In the latter case, the trend may be explained by correlation with properties such as \textbf{productivity}, with expressions in lower productivity ranges obtaining better representations.
More generally, the inconsistencies may stem from the use of different datasets, task formulations, and modeling approaches. This highlights the need for more systematic investigations of frequency effects.

Due to  \textbf{cross-linguistic variability} in MWE realizations, their optimal computational representations may differ across languages.
Yet our ability to assess these trends is limited: most surveyed work focuses on English, with one to two papers analyzing Chinese, Gallician, German, Japanese, Portuguese, Russian, and Slovenian.
We have not identified language-specific patterns for comparable experiments conducted on different languages.

Beyond our survey perspective, direct evidence of cross-linguistic variability is provided by limited studies on two languages in parallel.
Some of these report broadly comparable cross-linguistic patterns, e.g.\ on probing for compound semantics in English and Portuguese \citep{garcia2021probing}
and on idiom token classification in English and Japanese \citep{takahashi2022leveraging}.
Varied cross-linguistic differences have been observed elsewhere.
Comparable phrase similarity experiments have obtained the best results using sentence embeddings for English and verb embeddings for Galician \citep{gamallo2021comparing, gamallo2022evaluating}.
On compound compositionality prediction, the best model for English is (monolingual) BERT, and for Portuguese (multilingual) SBERT \citep{garcia2021assessing}.
A PIE identification experiment has found better monolingual generalizability for English than Russian, and better cross-lingual transfer from Russian to English than vice-versa \citep{fakharian2021contextualized}.
But each of these cases is a single point of reference, making it unclear if the trends are due to model or dataset properties rather than cross-linguistic differences.

\textbf{Summary.} 
Beyond inherently more informative MWEs, transformer representations are better with lower semantic idiosyncrasy and dispersion (cf.\ polysemy, productivity).
They also appear to be biased towards concrete expressions, while
the precise effect of other factors such as frequency remains unclear.
Cross-linguistic analyses are limited -- both regarding the coverage of different languages and direct comparisons across them -- with further work needed to establish reliable trends.

\section{Conclusion and outlook}
\label{sec:conclusion}
We have presented a survey of recent work on MWE semantics in transformer-based language models.
Starting with a general assessment of pretrained representations, we have seen that they capture some aspects of MWE meaning, but this ability is neither comprehensive nor consistent.
It can in principle be improved with optimization strategies such as fine-tuning and knowledge enhancement, but with highly variable gains.
MWE representations rely on memorized information rather than sophisticated meaning processing; this is reflected by suboptimal generalization ability.

Turning to differences in representational information in model architecture and textual input, we find that the most adequate representations are those corresponding to the linguistic structure of interest modeled within broader context; this enables the attention mechanism to efficiently encode expression-level information.
There is also broad consensus that lower layers are better at capturing MWE meaning, as observed on tasks such as predicting compound compositionality, PIE idiomaticity, and paraphrase similarity.
However, implementation decisions should always be carefully tuned because of interactions with other factors.
This includes a range of linguistic properties, with better representations for expressions exhibiting less semantic idiosyncrasy and dispersion.

The surveyed papers provide varied and valuable insights, but many conclusions are not directly comparable and cannot be extrapolated across MWE types or models.
We particularly underscored this issue regarding (i)~optimization strategies, which may interact with target expression types, models, and evaluation tasks; (ii)~layer-wise processing mechanisms, with similar patterns for memorized and compositional expressions; and (iii)~cross-linguistic variability, with insufficient evidence to identify broad trends.

Future studies can address these challenges through several lines of work.
(i)~Extending the coverage of MWEs beyond the current focus on compounds and idioms, ideally in a comparative setup, and systematically accounting for the effect of their linguistic properties.
(ii)~Extending the coverage of non-English languages, including in cross-linguistic evaluations.
(iii)~Broadening the scale of evaluations by multiplying experimental parameters, e.g.\ by investigating model structures across architectures as well as MWE types.
(iv)~Formulating tasks that are challenging in terms of both core semantic mechanisms and generalization requirements.
We believe that these perspectives will help disentangle interactions between experimental parameters and improve the generalizability of the resulting claims.

\section*{Acknowledgements}
We thank Michael Roth as well as the action editor and anonymous reviewers for valuable feedback.
This research was supported by the DFG Research Grant SCHU 2580/5-1 (\textit{Computational Models of the Emergence and Diachronic Change of Multi-Word Expression Meanings}).

\bibliography{mwe_survey, mwe_background}
\bibliographystyle{acl_natbib}
\end{document}